\documentclass[a4paper,conference]{IEEEtran}

\usepackage{graphics}    
\usepackage{times}       
\usepackage{amsmath}     
\usepackage{amssymb}     
\usepackage{graphicx}
\usepackage{algorithm}
\usepackage[noend]{algpseudocode}

\def\secref#1{Sec.~\ref{#1}}
\def\figref#1{Fig.~\ref{#1}}
\def\tabref#1{Tab.~\ref{#1}}
\def\eqref#1{Eq.~(\ref{#1})}

\newcommand\etal{\emph{et~al.}}

\def\argmin{\mathop{\rm argmin}}

\title{\Large Radar-based Automotive Localization using Landmarks in a Multimodal Sensor Graph-based Approach} 

\author{
\IEEEauthorblockN{Stefan J\"urgens}
\IEEEauthorblockA{
MAN Truck \& Bus SE\\
stefan.juergens@man.eu}
\and
\IEEEauthorblockN{Niklas Koch}
\IEEEauthorblockA{
Volkswagen AG\\
niklas.koch@volkswagen.de}
\and
\IEEEauthorblockN{Marc-Michael Meinecke}
\IEEEauthorblockA{
Volkswagen AG\\
marc-michael.meinecke@volkswagen.de}
}
\begin{document}
\maketitle
\thispagestyle{empty}
\pagestyle{empty}

\begin{abstract}
  %
	Highly automated driving functions currently often rely on a-priori knowledge from maps for planning and 
	prediction in complex scenarios like cities. This makes map-relative localization an essential skill. 

  In this paper, we address the problem of localization with automotive-grade radars, using a real-time 
	graph-based SLAM approach. 
  The system uses landmarks and odometry information as an abstraction layer. This way, besides radars, 
	all kind of different sensor modalities including cameras and lidars can contribute. 
	A single, semantic landmark map is used and maintained for all sensors. 
	
  We implemented our approach using C++ and thoroughly tested it on data obtained with our test vehicles, comprising
	cars and trucks. Test scenarios include inner cities and industrial areas like container terminals. 
	The experiments presented in  this paper suggest that the approach is able to provide a precise and stable 
	pose in structured environments, using radar data alone. The fusion of additional sensor information from cameras or lidars 
	further boost performance, providing reliable semantic information needed for automated mapping. 
\end{abstract}

\section{Introduction}
\label{sec:intro}

Automated driving functions are a current focus of automotive research activities. These modern systems require rich and redundant sensor information. Additionally, a-priori knowledge, taken from maps, is needed for planning and prediction in complex scenarios like inner cities or container terminals. This makes a precise, map-relative localization an essential prerequisite for L4 and L5 driving functions in such complex environments. 

Localization systems based on laser measurements, camera images and odometry information are well-known since decades. In contrast to this, radar-based odometry and localization are rather new techniques, under consideration of the research community since just a short time. While automotive-grade radars still suffer from a poor resolution compared to lidars, they provide undeniable advantages: reliable in harsh (weather-)conditions, tested and approved in the automotive mass market context for years, and thus quite cost-effective. 
In order to bring highly automated driving functions into series productions, these properties should be combined with the advantages 
of lidars and cameras to form a truly safe and redundant system.  

In this paper, we deal with the integration of radars into a graph-based SLAM (Simultaneous Localization and Mapping) 
system~\cite{grisetti2010titsmag}. Landmarks and odometry information are used as an abstraction layer, such that 
different types of sensors like radars, lidars, cameras and IMUs (inertial measurement units) can contribute to a single, joint minimization problem. 
This also offers the possibility for a single semantic landmark map for all sensors, making maintaining such a map 
in a large scale, series scenario feasible. We use a third-party, pre-recorded map, which we partly enriched 
using our SLAM system. 

\begin{figure}[t]
  \centering
 \includegraphics[width=1.0\linewidth]{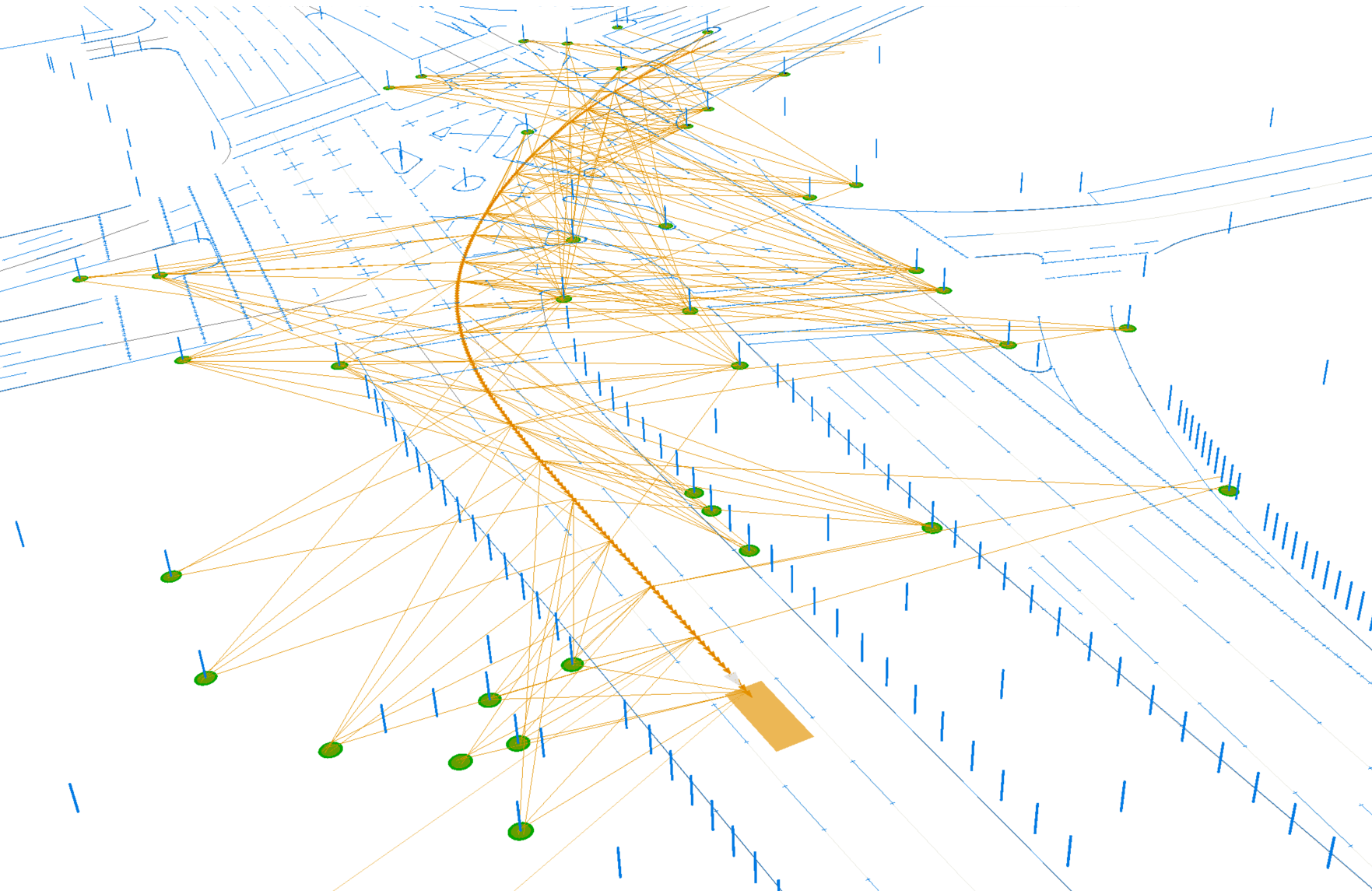}
 \caption{SLAM graph of Hamburg inner city, run on radar data alone. Map in blue, landmark measurements in orange, map associations in green and localization pose in orange. }
\label{fig:motivation}
\end{figure}

The main contribution of this paper is a graph-based SLAM system, that can run both on radar data alone 
as well as on a wide, multimodal range of sensors. See \figref{fig:motivation} for an example. 
We achieve this by the extraction of landmarks and odometry information out of the untracked radar reflection points of a single measurement cycle in real-time. This
allows us to provide a safe, redundant localization solution in real-world scenarios including inner cities and 
industrial areas like a container terminal. 

In sum, we make these key claims:
Our real-time SLAM approach is able to
\begin{itemize}
    \item (i) localize with \textbf{automotive-grade radar data alone} in challenging scenarios like inner city and container terminals.
		\item (ii) fuse information of \textbf{different sensor modalities}: 
		radar, lidar, camera, vehicle odometry and GNSS, using 
		a \textbf{single, semantic landmark map} for all sensors.
\end{itemize}
These claims are backed up by the paper and our experimental evaluation in~\secref{sec:exp}.

\section{Related Work}
\label{sec:related}

There is already some literature on radar-based localization and its sub-problems. We present a short summary, divided into two main areas:

\subsection{Radar-based odometry}
The radars on a vehicle measure the radial, relative velocity of the surrounding targets via the Doppler effect. Assuming that a good part of the environment is 
static, one can calculate the movement of the vehicle out of the Doppler velocities of these static targets. This is done in 
Kellner~\etal~\cite{kellner2014icra,kellner2016phd}, putting all measurements in a large set of equations and solving them using different Least-Squares estimators (LSQ). Our own 
radar-based odometry follows this approach very closely, with a few additions. 

Barjenbruch~\etal~\cite{barjenbruch2015iv} use a different, density-based framework. Here, measurements of Doppler radar sensors 
are represented as a mixture of Gaussian distributions. The ego-motion of the vehicle is obtained by calculating the transformation between these Gaussians 
of subsequent measurements. Both Doppler velocity and spatial information is used here. A variant with significant lower computational cost
is presented in Rapp~\etal~\cite{rapp2015ecmr}. 
As we want the spatial tracking to be solely on the landmark level, we do not follow this approach. 

\subsection{Localization and Mapping}
Ward and Folkesson~\cite{ward2016iv} consider localization wrt. to a pre-recorded map. Two short range radars (SRR) provide point detections, 
which are registered to a map in every time step using the Iterative Closest Point algorithm (ICP). An 
Extended Kalman Filter (EKF) provides the map-relative vehicle pose. 

Hammarsten and Runemalm~\cite{hammarsten2016master} consider radars which provide measurements in both azimuth and elevation. They are used 
to construct 3D occupancy grid maps. Together with inertial measurements from the vehicle, they calculate a full 6D pose using two different methods: 
a particle filter and a registration-based algorithm. 

In \cite{lupfer2017icmim}, Lupfer~\etal~show that the FastSLAM~2.0 algorithm benefits from the inclusion of radar data. 
They extend the measurement model for the landmarks to include, besides the usual angle and distance, also the Doppler information. Additionally to the 
radar landmarks, the vehicle odometry is used in the algorithm. 

The approach by Schuster~\etal~\cite{schuster2016itsc} describes a complete, graph-based SLAM framework using 4 radars mounted on a car, providing 
$360^{\circ}$ coverage. Point landmarks with unique descriptors are extracted from the data, associated to a map, and solved, together with an odometry 
from an IMU and wheel encoders, in a graph. This has similarities to our approach, but uses a wheel-based odometry and point landmarks that 
are specific for radars and do not necessarily correspond to semantic objects. 

\section{Our Approach}
\label{sec:main}

In this paper, we present a graph-based SLAM framework using landmark, odometry and global information. It has 
the following design goals:
\begin{itemize}
    \item Fusion of information of \textbf{different sensor modalities}: 
		radar, lidar, camera, odometry e.g. from inertial sensors and wheel encoders, and GNSS. 
    \item Use of a \textbf{single, semantic landmark map} for all sensor modalities.  
		It can be constructed by our system or pre-recorded by an external supplier. 
		\item \textbf{Modularity}: Split of the system into the three parts: landmark extraction, odometry calculation 
		and graph construction and minimization. 
		\item \textbf{Sensor redundancy}: The system should perform stably with only a subset of possible sensor inputs, e.g. radar-only or lidar and odometry information
		alone. 
\end{itemize}

The used graph-based approach was already presented in~\cite{wilbers2019icra,wilbers2019irc-amws}, mainly run on lidar, odometry and GNSS data. We recap it shortly 
in~\secref{subsec:graph_theory}. 
The new feature presented in this paper is that we can run the SLAM algorithm on radar data alone. The calculation of the needed
radar-based odometry is detailed in \secref{subsec:odometry_theory}, the landmark extraction is presented in \secref{subsec:radar_feature_theory}. 

\subsection{Graph-based SLAM}
\label{subsec:graph_theory}
In our landmark-based SLAM, we want to calculate the poses of the vehicle and landmarks that best fit to both the sensor measurements made in the past as well as to a pre-recorded map, if already existing. Measurements can be of the three types: landmark position, odometry measurement or global pose measurement (e.g. GNSS). 

Writing this as a minimization problem, we define the state $\boldsymbol{x}=[\boldsymbol{x}^p \  \boldsymbol{x}^l]$, with the vehicle poses $\boldsymbol{x}^p=[\boldsymbol{x}_1^p,\dots,\boldsymbol{x}_P^p]$ and landmark positions $\boldsymbol{x}^l=[\boldsymbol{x}_1^l,\dots,\boldsymbol{x}_L^l]$. The number of poses $P$ determines the length of our sliding window, and $L$ is the number of landmark estimates. Then the task is to solve  
\begin{align}
\label{eq:graph_minimize}
  \boldsymbol{x}^* &= 
	 \argmin_{\boldsymbol{x}} \sum_i \boldsymbol{e}_i (\boldsymbol{x},\boldsymbol{z}_i)^T \boldsymbol{\Omega}_i \boldsymbol{e}_i (\boldsymbol{x},\boldsymbol{z}_i)
	+ \mathcal{F}^{map}(\boldsymbol{x}^l). 
\end{align}
Here, we introduce the error functions $\boldsymbol{e}_i$ and the information matrices $\boldsymbol{\Omega}_i$ related to the measurements $\boldsymbol{z}_i$. Both incorporate the information of the measurements and made associations with the map. They differ for the different measurement types of odometry, global poses, landmark observations and landmark map associations. For the latter, we can specify
\begin{align}
\label{eq:graph_map_errors}
  \mathcal{F}^{map}(\boldsymbol{x}^l) &= 
	\sum_i \boldsymbol{e}_i^{map} (\boldsymbol{x}^l)^T \Omega_i^{map} \boldsymbol{e}_i^{map} (\boldsymbol{x}^l)
\end{align}
with the error function $\boldsymbol{e}_i^{map} (\boldsymbol{x}^l) = \boldsymbol{x}_i^l - \boldsymbol{m}_i$. The global positions of the map 
landmarks are given by $\boldsymbol{m}_i$. 

This minimization problem can also be viewed as a factor graph: here, the state variables correspond to nodes, while measurements, error functions and information matrices can be seen as factors. 

Our algorithm has three main steps. First in \textbf{local association}, new landmark measurements are associated to landmark states. These can be added to the graph. In the second step \textbf{map association}, landmark states are associated to the landmarks of the map. In the last \textbf{optimization} step, the problem is solved numerically. The library g2o~\cite{grisetti2010titsmag} is used here. 

As landmarks, we use static, semantic objects that can be detected by different sensor modalities like radar, lidar and camera. This enables us to use a single, globally referenced landmark map for all sensors. Additionally, the landmark density should be sufficiently high in all considered scenarios. Examples are pole-like objects, building corners and planes, lane markings, curbs and guardrails. 

For automotive-grade radars, this can lead to problems. Due to the 2D nature of the point data and the low point density, the extraction of semantic information is very difficult. Effectively, one can only extract point-like and line-like features out of the radar data. Associating these to semantic landmarks is quite unreliable. Using unique descriptors as in~\cite{schuster2016itsc} could help, but they would need to be stored in the landmark map as well. We decided for a simpler approach, where we associate radar point features with poles, and radar line features with planes, curbs and guardrails. While for mapping this is quite problematic, for localization it works surprisingly well. 

\subsection{Radar-based odometry}
\label{subsec:odometry_theory}
For the calculation of the odometry from the radar data, we closely follow the approach by Kellner~\etal~\cite{kellner2014icra,kellner2016phd}. 
Here, we present a short summary, including some tweaks and additions we made. For details, we refer to the original publications. 

\begin{figure}[t]
  \centering
  \includegraphics[width=0.99\columnwidth]{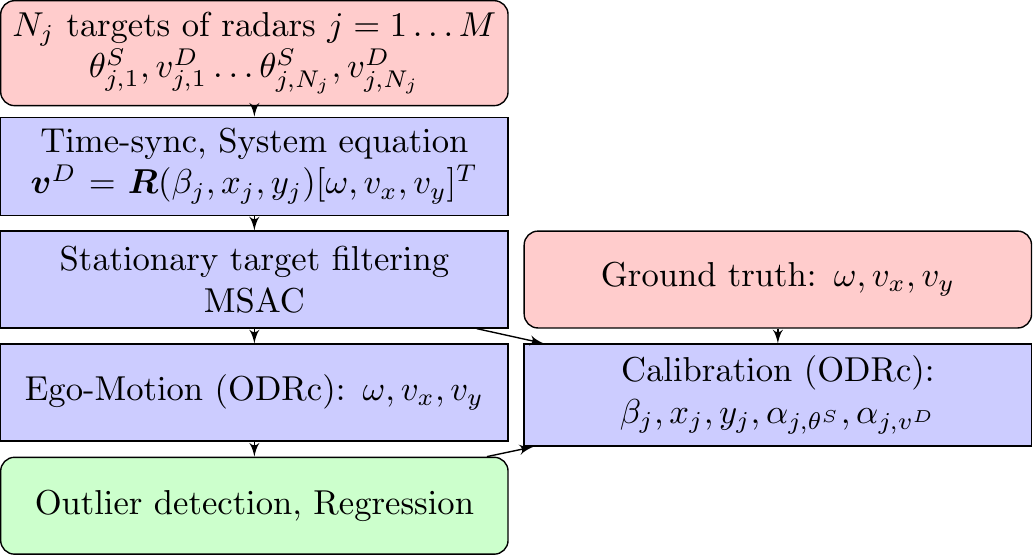}
  \caption{The main steps of the radar-based odometry algorithm. }
  \label{fig:radarodo_algorithm}
\end{figure}

The basic idea of the approach is to use the measured Doppler velocity of the targets around the vehicle to 
obtain the ego-motion of the vehicle. This is possible as a large amount of targets is usually static, hence the 
Doppler velocities of these targets stem from the motion of the ego-vehicle. 
The algorithm flow is depicted in \figref{fig:radarodo_algorithm} and explained in the following. 

In \textbf{Step 1}, the $N_j$ untracked reflection points from different radars $j=1,\dots,M$ mounted on the vehicle are collected over a period of time, with Doppler velocity $v_{j,i}^D$, azimuth angle $\theta_{j,i}^S$, 
timestamp $T_j$ and $i\in\{1, \dots, N_j\}$. They are put into a single 
system of equations, relating them to the vehicle's yaw rate $\omega$ and velocity $v_x$ and $v_y$: 
\begin{align}
\label{eq:odo_overall}
	\begin{bmatrix}
	  \alpha_{1,v^D} \boldsymbol{v}_{1}^{D} \\
		\vdots \\
		\alpha_{M,v^D} \boldsymbol{v}_{M}^{D}
	\end{bmatrix} &= 
	\begin{bmatrix}
		\boldsymbol{M}_{1} \boldsymbol{S}_{1} \\
		\vdots \\
		\boldsymbol{M}_{M} \boldsymbol{S}_{M}
	\end{bmatrix} 
	\begin{bmatrix}
		\omega \\
		v_x \\
		v_y
	\end{bmatrix}.
\end{align}
with $\boldsymbol{v}_{j}^{D} = [v_{j,1}^D, \dots, v_{j,N_j}^D]^T$. The matrices
\begin{align}
\label{eq:s_j_m_j}
  \boldsymbol{S}_{j} &= 
	\begin{bmatrix}
		-y_j & 1 & 0 \\
		x_j & 0 & 1 
	\end{bmatrix} \\
	\boldsymbol{M}_{j} &= 
	\begin{bmatrix}
		\cos(\theta_{j,1}) & \sin(\theta_{j,1}) \\
		\vdots & \vdots \\
		\cos(\theta_{j,N_j}) & \sin(\theta_{j,N_j})
	\end{bmatrix}
\end{align}
with $\theta_{j,i} = \beta_j + \alpha_{j,\theta^S} \theta_{j,i}^S$ contain the mounting angle $\beta_j$ and position $(x_j, y_j)$ of the radar sensors mounted on the vehicle. $\alpha_{j,v^D}$ and $\alpha_{j,\theta^S}$ are scaling factors for every sensor added by us in order to deal 
with bias errors in measured Doppler velocity and azimuth angle. They are determined by our calibration procedure. 

As an addition, we consider the time-sync of the data. The point data from different radars was measured at different times $T_j$, thus one should synchronize them to a common timestamp $T_{odo}$. The latter is ideally the 
middle of the measurement window, minimizing bias. This is done by replacing in \eqref{eq:odo_overall} 
$\boldsymbol{v}_{j}^{D} \Rightarrow \boldsymbol{v}_{j}^{D} - \boldsymbol{\Delta v}_{j}$ and 
$\boldsymbol{M}_{j} \Rightarrow \boldsymbol{M}_{j}^{T_{odo}}$ with 
\begin{align}
	\boldsymbol{\Delta v}_{j} &= \boldsymbol{M}_{j} \boldsymbol{S}_{j} 
	\begin{bmatrix}
		\omega^{T_j} \\
		\boldsymbol{v}^{T_j}
	\end{bmatrix} 
	- \boldsymbol{M}_{j}^{T_{odo}} \boldsymbol{S}_{j} 
	\begin{bmatrix}
		\omega^{T_{odo}} \\
		\boldsymbol{v}^{T_{odo}}
	\end{bmatrix} \\ 
	\boldsymbol{M}_{j}^{T_{odo}} &= 
	\begin{bmatrix}
		\cos(\theta_{j,1} + \Delta\psi_j) & \sin(\theta_{j,1} + \Delta\psi_j) \\
		\vdots & \vdots \\
		\cos(\theta_{j,N_j} + \Delta\psi_j) & \sin(\theta_{j,N_j} + \Delta\psi_j)
	\end{bmatrix} \\
	\Delta\psi_j &= \psi^{T_j} - \psi^{T_{odo}}. 
\end{align}
The needed rates $(\omega^{\tau}, \boldsymbol{v}^{\tau})^T$ and vehicle headings $\psi^{\tau}$ at different times $\tau \in \{T_j, T_{odo} \}$ can be obtained by extrapolation of the results of 
the last cycles of the odometry calculation. 

In \textbf{Step 2}, a two stage filtering is applied to remove all targets from the equation which stem from moving objects in the vicinity of the ego-vehicle. At first, all measured velocities are compared to the ego-motion results of the last cycles and unreasonable data is thrown out, based on a simple vehicle model. 
Afterwards, the random consensus method MSAC~\cite{torr00} is used to retain only the static targets, which are expected to form the majority of the targets in the system of equations. As we want to obtain the 3~DOFs $\omega$, $v_x$ and $v_y$, this corresponds to fitting the data to a 3-dimensional plain. As $v_y$ is usually small in most scenarios, we prefer for simplicity and lower run-time to substract $v_y$ from the equations, based on previous ego-motion results of the algorithm, and use MSAC together with a normal 2D plane. 
We choose the threshold distance $d_{MSAC}$ of the data points to the plane based on the standard deviations $\sigma_{v^D}$ and $\sigma_{\theta^S}$ of the measured variables $v_{j,i}^D$ and $\theta_{j,i}^S$, which is provided by the radar manufacturer. Choosing $d_{MSAC}$ such that between $60\%$ and $80\%$ of the static measurements are used leads to the best results for us. 

In \textbf{Step 3a} we obtain the ego-motion solution for $(\omega, v_x, v_y)^T$, using the orthogonal distance regression algorithm with bias compensation (ODRc). 

In \textbf{Step 4a}, the obtained ego-motion is compared to the previous results to detect outliers. In these failure cases, we perform a regression over the previous ego-motion results to obtain a best guess for the vehicle state at the present timestamp. 

The ego-motion result is very sensitive to deviations in the calibration of the yaw angle of the mounted sensors. We therefore implemented a second path (\textbf{Step 3b}) of the algorithm, used for calibration. Here, we take the ego-motion result $(\omega, v_x, v_y)^T$, either from a reference system or from the radar odometry itself, and calculate the calibration parameters $(\beta_j, x_j, y_j, \alpha_{j,v^D}, \alpha_{j,\theta^S})^T$ for every radar separately. In the online configuration, the parameters are calculated every cycle and collected in a median filter to obtain the most likely calibration parameters after a longer period of time, e.g. a drive of $10\,\mathrm{min}$. In the offline version, the measurements of the complete drive, minus outliers, are collected and optimized as a whole. This leads to better calibration parameters, especially for the sensor mounting positions which can only be reliably determined in driving situations with a finite yaw rate. 

The algorithm works very precisely and reliably, if enough static targets are available. A $360^{\circ}$ coverage of the surroundings of the vehicle is thus recommended. This way, the algorithm remains stable also in difficult situations like dense traffic, as shown in \secref{subsec:radar_robust}. 

\subsection{Radar feature detection}
\label{subsec:radar_feature_theory}
Besides the odometry information, the graph needs landmark measurements to estimate a map relative pose. The main 
problem here is that our radar data is noisy and only 2D, see~\secref{subsec:sensor_setup}. This makes it hard to extract semantic information about 
the measured objects, which would be necessary to distinguish landmarks of specific type. 
We therefore restrict ourselves to extract point-like and line-like features out of the measured radar data. These are 
then associated to landmarks as described in~\secref{subsec:graph_theory}. 

\textbf{Step 1}:
The untracked reflection point data of all radars is aggregated for some time, in our case $400\,\mathrm{ms}$. Using the 
Doppler velocity and odometry information, all points from moving objects are filtered out.  

\textbf{Step 2}:
We find point-like clusters in the data using the density-based clusterer OPTICS~\cite{ankerst99optics}. It is 
well-suited for the detection of clusters with different densities, which occur due to the different reflection 
point density of the radar data. All reflection points belonging to a point-like feature are removed from the data set. 

\textbf{Step 3}:
We search for line-like feature in the remaining set of points. MSAC~\cite{torr00} is used for this. Optionally, overlapping line segments can be merged.  

\section{Experimental Evaluation}
\label{sec:exp}

The experiments are designed to show the capabilities of our method and to
support our key claims, that our real-time SLAM approach is able to
(i) localize with \textbf{automotive-grade radar data alone} and 
(ii) fuse information of the \textbf{different sensor modalities} radar, lidar, camera, vehicle odometry and 
GNSS, using a \textbf{single, semantic landmark map} for all sensors.

All experiments are performed with our car or truck test vehicles presented in ~\secref{subsec:sensor_setup}. 
The driven scenarios are presented in~\secref{subsec:test_scenarios}. The overall localization performance of the full system is evaluated in ~\secref{subsec:loc_exp}
on this real world data against a reference system. 
In \secref{subsec:radar_robust}, we have  a closer look at the stability and precision of the radar-based odometry, deducing necessary, scenario dependent criteria under which the performance of the odometry, and the whole localization system running on radar data alone, remains stable. 

\subsection{Test Vehicles: Car and Truck}
\label{subsec:sensor_setup}
\begin{figure}[t]
  \centering
  \includegraphics[width=0.676\columnwidth]{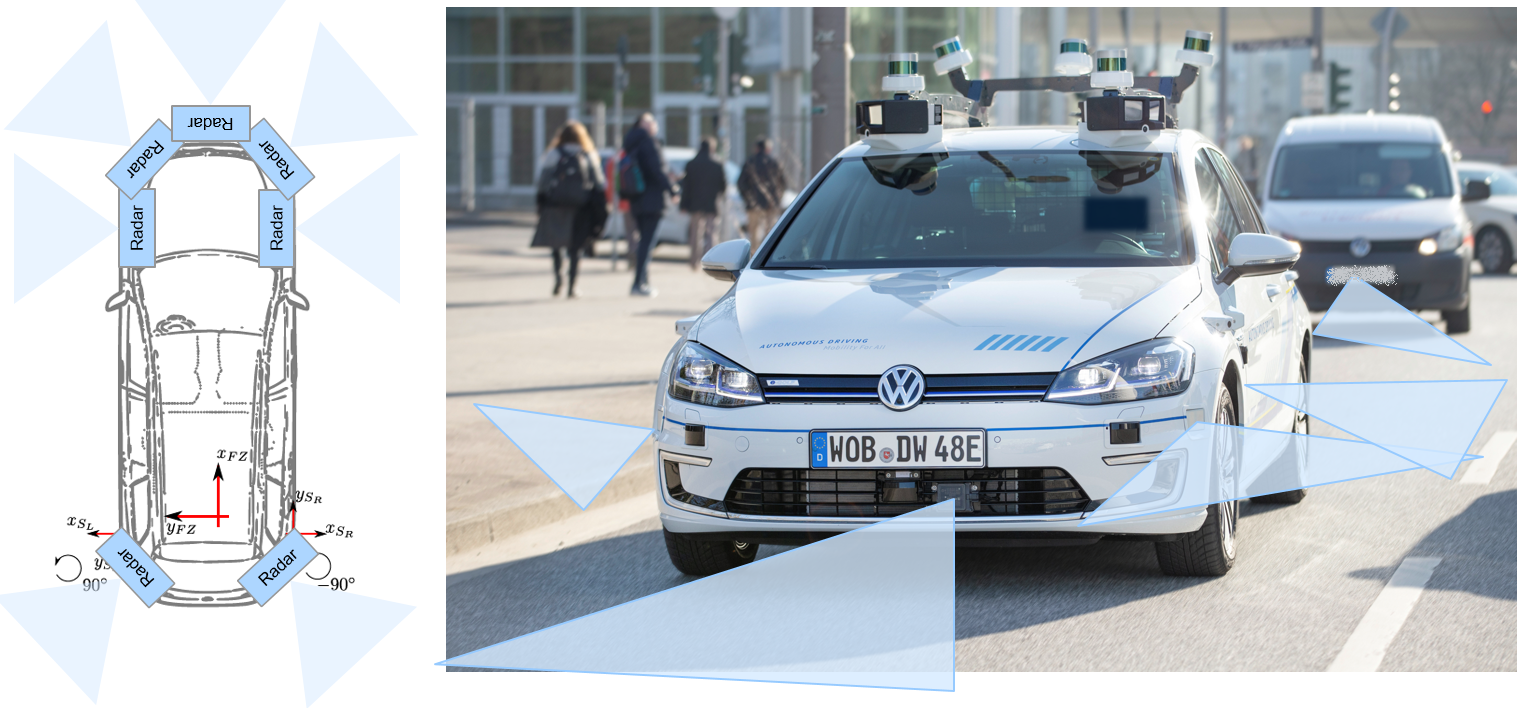}
	\includegraphics[width=0.31\columnwidth]{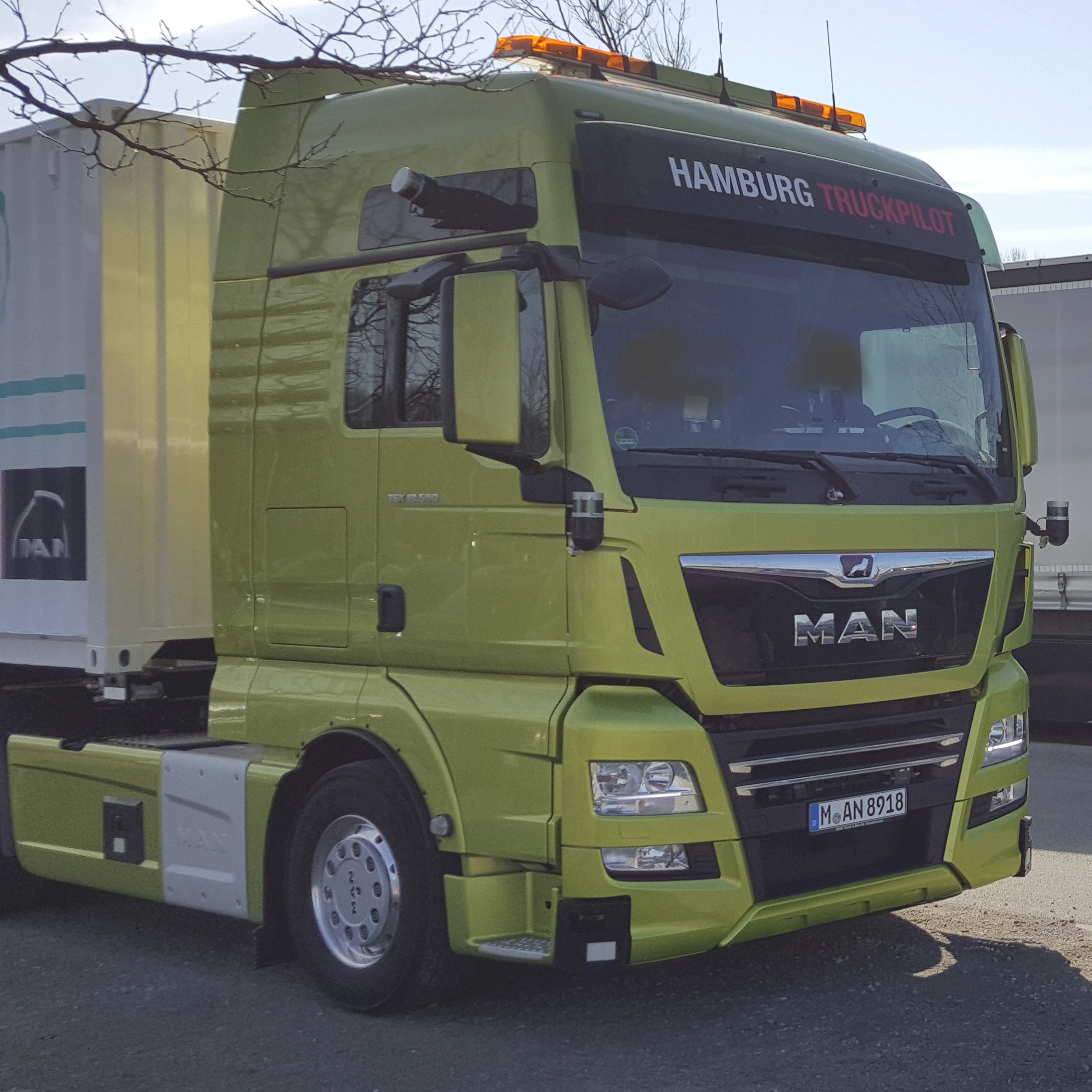}
  \caption{Sensor setup of our vehicles. The cars use the 6 radars on the corners and on the sides, the trucks currently only 4: the two on the sides and on the front corners, respectively. }
  \label{fig:sensor_setup}
\end{figure}

We perform our experiments on two different test vehicles: an e-Golf~7 and a MAN semi-trailer truck, see \figref{fig:sensor_setup}.
While the vehicles look and drive quite differently, the sensor setup is similar. They use wheel encoders and an 
IMU for the classical vehicle odometry, a close-to-series GNSS receiver, Velodyne lidars, a front camera for lane detection, and have a reference-grade 
RTK system installed: for the truck it's an ADMA, for the car an Applanix. Both use $77\,\mathrm{GHz}$ automative-grade radars: While the cars 
have a $360^{\circ}$ coverage by 6 short range radars (SRR), the trucks 
only use 4 of the SRRs. An additional longe range radar in the front is currently not used in our experiments. 
The radars are installed at a height of 40-70\,cm which is not optimal for localization purposes as their line of sight to landmarks is often blocked by other traffic participants. The data rate is $20\,\mathrm{Hz}$, which we also choose as the output frequency of the odometry. 

Corominas-Murtra~\etal~\cite{corominas2016icra} show that for odometry calculations, 
pairs of radars mounted diagonally on the vehicle perform best. Therefore we expect the truck setup 
to be less optimal than the car setup. In general, our current automotive-grade radars provide only 2D point clouds 
and suffer from poor resolution. New developments like synthetic aperture radars (SARs) discussed by Gisder~\etal~\cite{gisder2018irs,gisder2019irs}  
or polarimetric radars presented in~\cite{weishaupt2019irs} are expected to improve the data quality significantly in the future. 

\subsection{Test Scenarios: Inner City and Container Terminal}
\label{subsec:test_scenarios}
Our test scenarios cover two of the relevant and complex cases for cars and trucks: \textbf{Hamburg inner city} and a \textbf{container terminal} in Hamburg harbor. During the drive, traffic was quite dense (moderate) in the city (terminal), respectively. 
Third-party, pre-recorded maps based on high dense lidar data, that were initially meant for lidar and camera-based localization, not radars, are used here. This works fine, although the radar sees the world a bit differently compared to a lidar, e.g. detecting the poles of a guardrail instead of the rail itself. 

\begin{figure}[t]
  \centering
  \includegraphics[width=0.8\columnwidth]{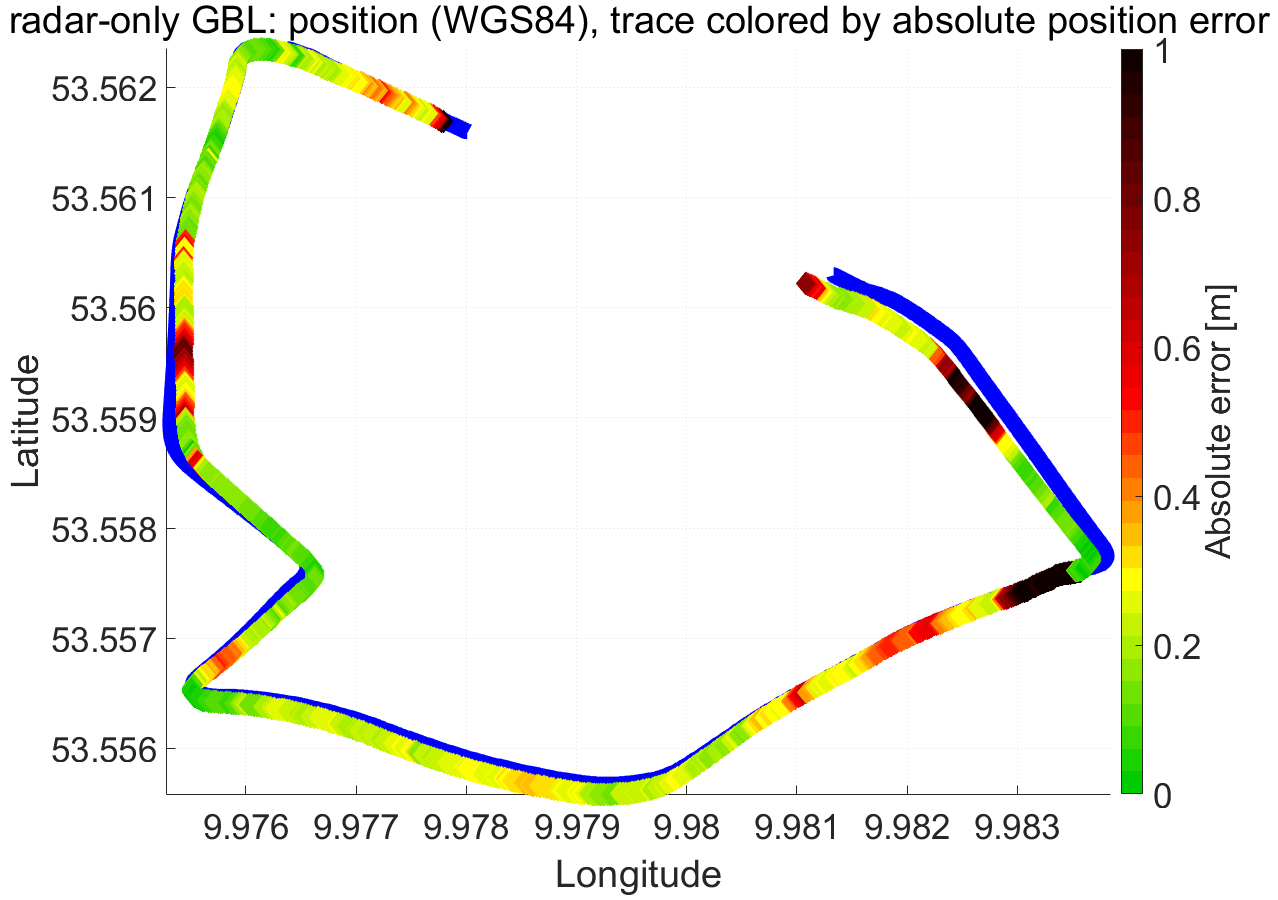}
  \includegraphics[width=0.8\columnwidth]{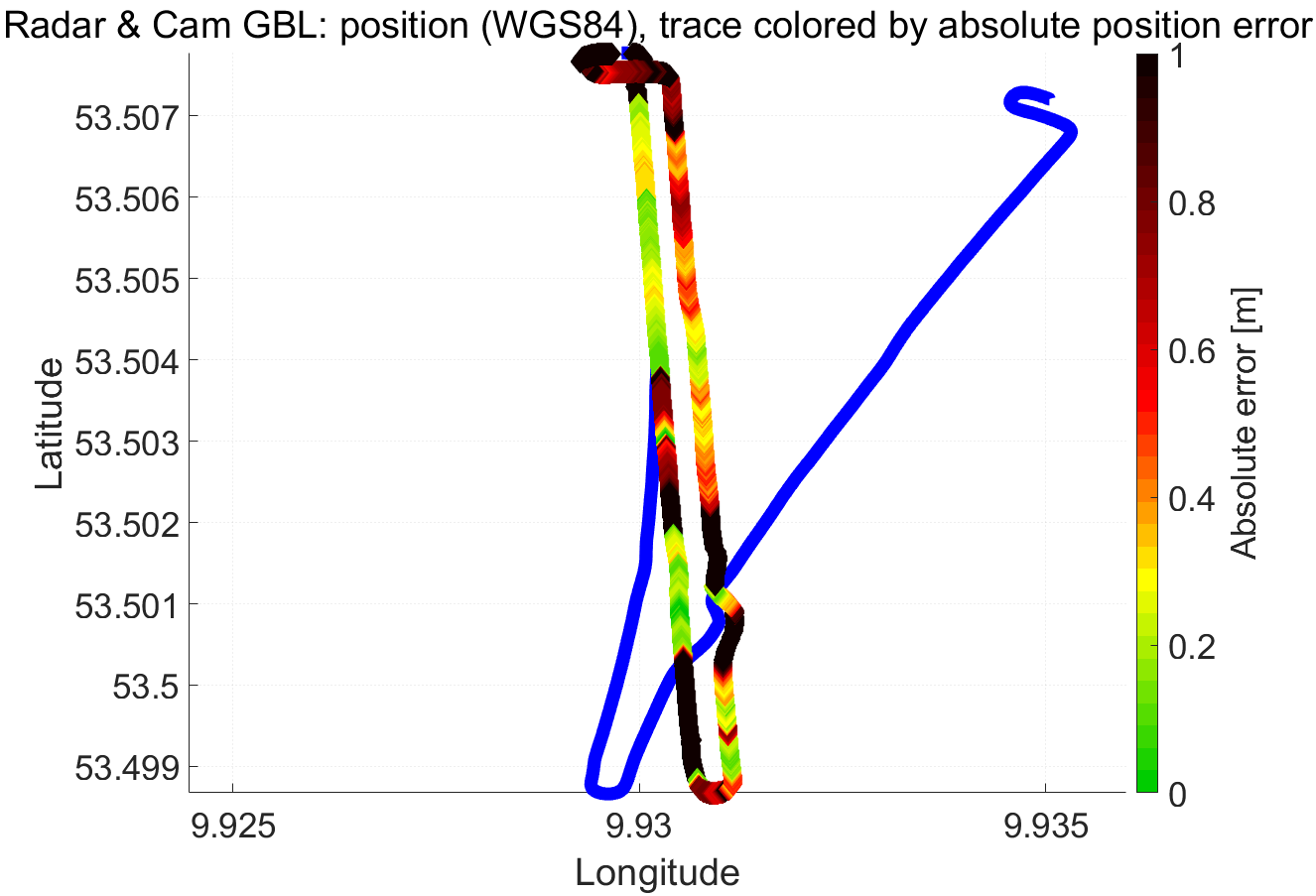}
  \caption{Driven paths for the city (top) and terminal (bottom) scenario. The colors indicate the absolute position error 
	of the radar-only (top) and radar\,\&\,camera (bottom) graph-based localization (GBL) wrt. the reference system. Additionally, the path of the radar odometry is plotted in blue. }
  \label{fig:loc_path}
\end{figure}

In the experiments, we compare the driven paths for radar odometry and the localization system in the three cases radar-only, 
radar\,\&\,camera localization, and the full sensor setup including the lidars and the vehicle odometry. The RTK reference systems of the vehicles serve as ground truth. The driven paths are visualized in \figref{fig:loc_path}, 
where the color coding indicates the absolute position error. The paths obtained by the radar odometry alone are shown in blue. 

\subsection{Overall Localization Accuracy}
\label{subsec:loc_exp}
\begin{figure}[t]
  \centering
  \includegraphics[width=0.493\columnwidth]{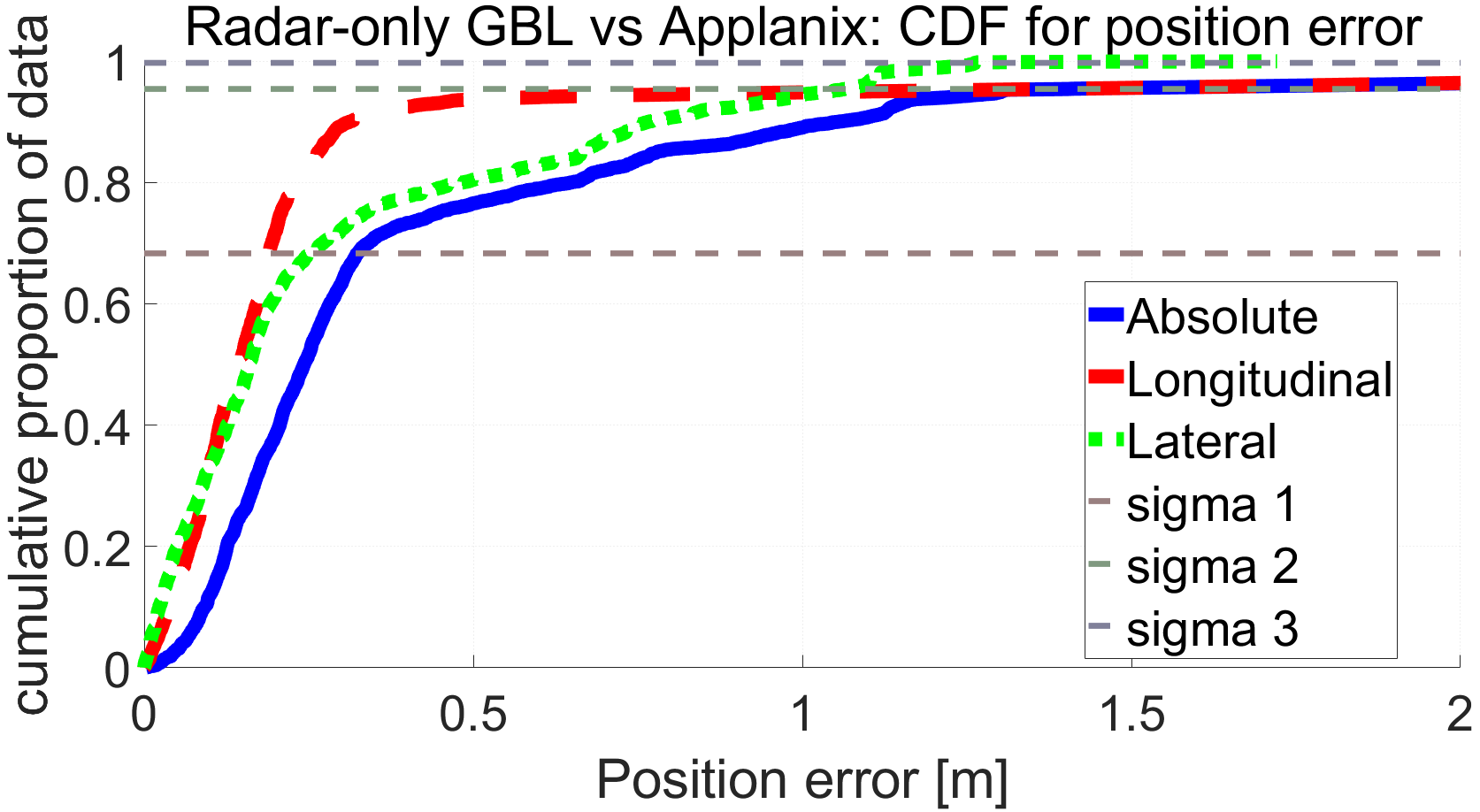}
  \includegraphics[width=0.493\columnwidth]{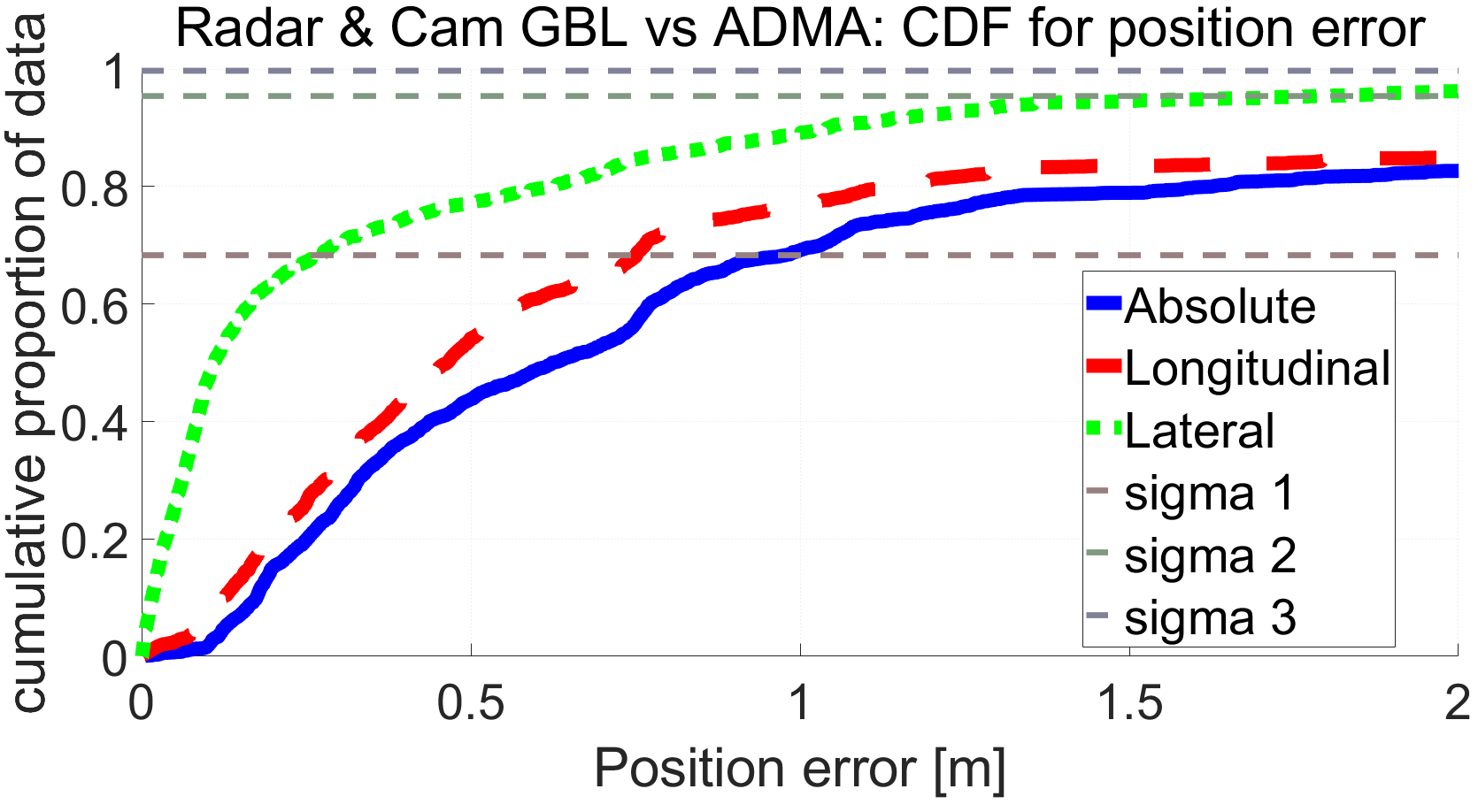}
  \caption{The CDF plots of the radar-only graph-based localization (GBL) for the city (left) and the radar\,\&\,camera localization for the terminal (right) scenario. }
  \label{fig:loc_accuracy}
\end{figure}

The standard deviation of the lateral and absolute position error wrt. the reference is shown in \tabref{tab:loc_accuracy}. 
\begin{table}
	\caption{Standard deviation of the lateral and absolute position error wrt. the reference. For $1\sigma$ ($2\sigma$), 68.2\% (95.4\%) of 
	the errors are below the given value.}
	\centering
	\begin{tabular}{|c|c|c|c|c|}
	\hline
													& $1\sigma$ city & $2\sigma$ city & $1\sigma$ termi. & $2\sigma$ termi. \\
	\hline
	radar-only lat.					& 0.25\,m & 1.07\,m & 0.75\,m & $>$2\,m \\
	radar-only abs.					& 0.32\,m & 1.41\,m & 1.16\,m & $>$2\,m \\
	\hline
	radar\,\&\,camera lat. 	& 0.13\,m & 0.76\,m & 0.27\,m & 1.79\,m \\
	radar\,\&\,camera abs. 	& 0.20\,m & $>$2\,m & 0.98\,m & $>$2\,m \\
	\hline
	full~setup lat.			  	& 0.07\,m & 0.18\,m & 0.17\,m & 0.28\,m \\
	full~setup abs.			  	& 0.15\,m & 0.28\,m & 0.24\,m & 0.45\,m \\
	\hline
	\end{tabular}
	\label{tab:loc_accuracy}
\end{table}
Plots of the cumulative distribution function (CDF) for the radar-only and radar\,\&\,camera case are presented in \figref{fig:loc_accuracy}. 

\emph{Radar only: }
In the city scenario, radar alone performs very well due to the abundance of static infrastructure. This leads to 
an accurate radar-based odometry and good landmark detections. Inaccuracies arise if the line of sight to the landmark is 
blocked by other static objects like parking cars. This can lead to wrong map association, which happened in the two areas in 
\figref{fig:loc_path} (up) with the error being larger than 1\,m. 

The terminal is more difficult. The odometry suffers from a drift in some areas, which will be discussed in \secref{subsec:radar_robust} in more details. 
Additionally, parked containers block the line of sight to the mapped static infrastructure, and 
are regularly mistaken for buildings. In these cases, the localization accuracy deteriorates to the level of the 
GNSS receiver or even below. Therefore, the 4 radars of our sensor setup alone are not enough for a stable localization in this scenario. 

\emph{Radar and camera: }
The lane markings from the front camera increase the accuracy and stability of our localization approach. 
The latter stems from the higher landmark density and variety, such that wrong map association can be avoided. 
Yet, lane markings usually do not 
compensate longitudinal errors and can introduce additional errors and noise in complex areas like crossings. 

\emph{Lidar, radar, camera and vehicle odometry: }
Including the full sensor setup, our approach performs at least as good as post-processed RTK systems in the 
discussed scenarios, especially in the city. The determined $1\sigma$ ($2\sigma$) distances in the lower rows of \tabref{tab:loc_accuracy} give therefore the boundaries of the accuracy of our reference systems. The worse numbers for 
the terminal stem from the fact that we only post-processed the Applanix, but used the online pose of the ADMA. 

\subsection{Radar Odometry Accuracy and Robustness}
\label{subsec:radar_robust}
Aside from the landmark detections, the localization performance depends on the stability and accuracy of the used odometry. For 
our radar-based odometry, we identified two main problems occuring in our scenarios. 

\textbf{Drift due to vacancy: } In the case that there is an abundance of static environment on one side of the ego-vehicle and nothing on the other (or a moving object is filtered out there), the odometry tends to calculate a false yaw rate leading to a small drift to the free side. A possible reason might be small calibration errors which usually cancel each other out if all radars contribute equally to the odometry. Yet, as all radars are calibrated independently of each other, such an interdependency between them should not occur. 
Adding the scaling factors $\alpha_{j,v^D}$ and $\alpha_{j,\theta^S}$ (see \secref{subsec:odometry_theory}) reduced the problem significantly. Still, the sensor model is quite basic and should be extended. 
	
\begin{figure}[t]
  \centering
  \includegraphics[width=1.\columnwidth]{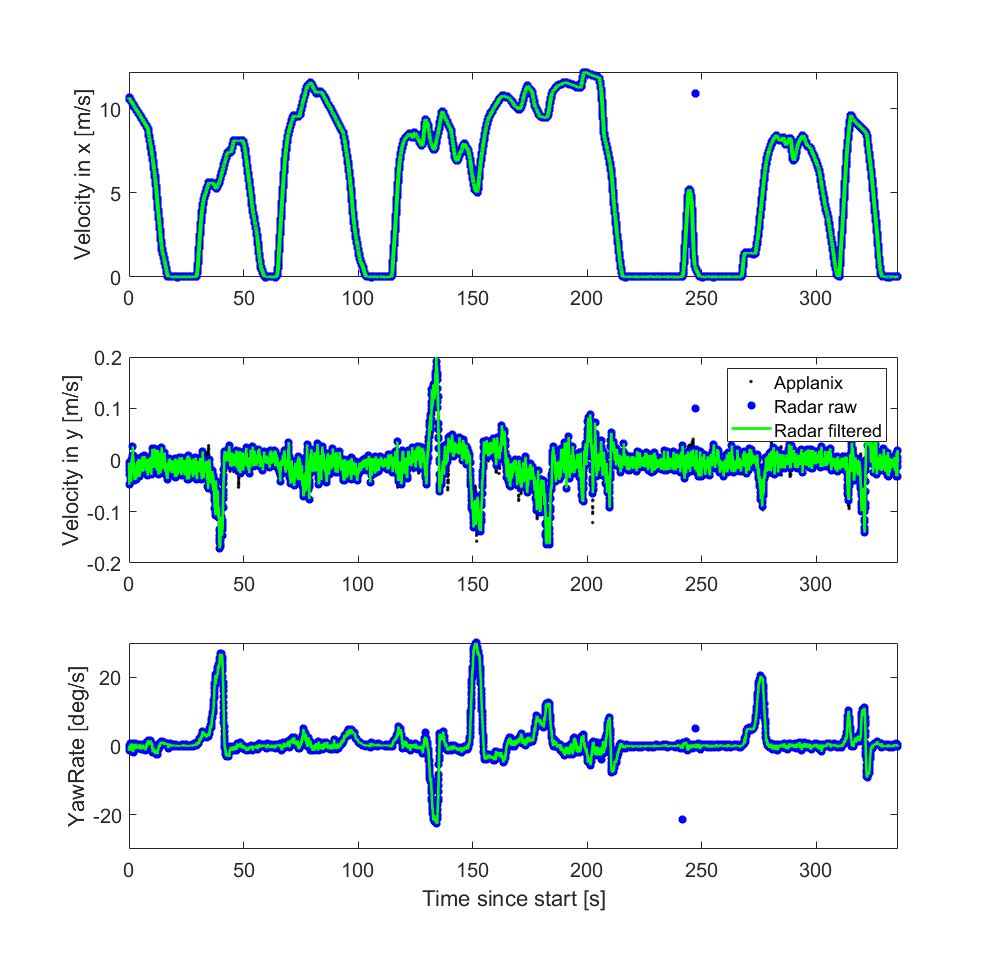}
  \caption{Yaw rate and velocities of the radar odometry for the Hamburg city scenario. 
	Radar raw (filtered) is the signal without (with) outlier rejection, respectively. }
  \label{fig:radar_odo_robust}
\end{figure}

\textbf{Drift / breakdown due to moving objects: } Dense traffic can lead to inaccuracies, if it is not filtered out completely. This occurs especially in stop-and-go situations with low ego-velocity. 
For the Hamburg city scenario this is only a minor problem, as indicated by the good odometry path in \figref{fig:loc_path}~(up). Additionally, we plot the velocities and yaw rate of the radar odometry in \figref{fig:radar_odo_robust}. 
During the whole drive the algorithm provided a stable solution, with the outlier detection and regression step 
being triggered only in two cycles (blue outliers in \figref{fig:radar_odo_robust}). The residuals have mean and standard 
deviation of
\begin{align}
\label{eq:residuals}
	r_{vx} &= 0.0046 \pm 0.019\,\mathrm{\frac{m}{s}}\\
	r_{vy} &= -0.0011 \pm 0.014\,\mathrm{\frac{m}{s}}\\
  r_{yawrate} &= -0.0052 \pm 0.40\,\mathrm{\frac{deg}{s}}.  
\end{align}
Thus the algorithms works very precisely and stably in dense traffic, which we attribute to the $360^{\circ}$ coverage of the radars. 

For the truck in the terminal, two situations of breakdown occured at the beginning and end of the drive: in both cases, a tractor-trailer crosses directly 
in front of our own truck, such that the majority of radar reflection points no longer stem from the static environment. 
As this crossing takes several seconds, the outlier detection and regression step can only partly cope with the problem. 
Additional radars that look at the back of the truck or are installed on the semi-trailer will easily fix this problem. 



\section{Conclusion}
\label{sec:conclusion}

In this paper, we presented a novel approach to use automotive-grade radar data for localization. 
Our method exploits a graph-based formulation using landmarks and odometry information, enabling a 
fusion of different sensor modalities while maintaining only a single, semantic landmark map. 
This allows us to successfully localize with radar data alone, as well as with an arbitrary combination of radar, lidar, and camera information. 

We implemented and evaluated our approach on different datasets taken with car and truck test vehicles. 
Our experiments suggest that radar-only localization works precisely and reliably in many scenarios, e.g. in inner cities. 
A fusion with additional sensor modalities like cameras 
and lidars can provide more detailed, semantic information, especially useful for mapping. 


\bibliographystyle{plain}

\bibliography{glorified,IRS2020_Localization}

\end{document}